\documentclass[letterpaper]{article} 
\usepackage{aaai2026}  
\usepackage{times}  
\usepackage{helvet}  
\usepackage{courier}  
\usepackage[hyphens]{url}  
\usepackage{graphicx} 
\usepackage{natbib}  
\usepackage{caption} 
\usepackage{multirow}
\usepackage{booktabs}
\usepackage{colortbl}
\usepackage{amssymb}
\usepackage{amsmath}
\usepackage{utfsym}
\usepackage{newfloat}
\usepackage{listings}
\DeclareCaptionStyle{ruled}{labelfont=normalfont,labelsep=colon,strut=off} 
\title{Distilling Future Temporal Knowledge with Masked Feature Reconstruction for 3D Object Detection}
\author {
    Haowen Zheng\textsuperscript{\rm 1}\thanks{Work done during the internship at HAOMO.AI Technology.},
    Hu Zhu\textsuperscript{\rm 2},
    Lu Deng\textsuperscript{\rm 3},
    Weihao Gu\textsuperscript{\rm 4}\equalcontrib,
    Yang Yang\textsuperscript{\rm 5}\equalcontrib,
    Yanyan Liang\textsuperscript{\rm 1}\equalcontrib
}
\affiliations {
    \textsuperscript{\rm 1}Macau University of Science and Technology\\
    \textsuperscript{\rm 2} The Hong Kong Polytechnic University\\
    \textsuperscript{\rm 3} HAOMO.AI Technology Co., Ltd.\\
    \textsuperscript{\rm 4} Institute for AI Industry Research, Tsinghua University\\
    \textsuperscript{\rm 5} MAIS, Institute of Automation, Chinese Academy of Sciences\\
    zhengnayin@gmail.com, guwh22@mails.tsinghua.edu.cn, yang.yang@nlpr.ia.ac.cn, yyliang@must.edu.mo
}

\begin{document}

\maketitle

\begin{abstract}
Camera-based temporal 3D object detection has shown impressive results in autonomous driving, with offline models improving accuracy by using future frames. Knowledge distillation (KD) can be an appealing framework for transferring rich information from offline models to online models. However, existing KD methods overlook future frames, as they mainly focus on spatial feature distillation under strict frame alignment or on temporal relational distillation, thereby making it challenging for online models to effectively learn future knowledge. To this end, we propose a sparse query-based approach, Future Temporal Knowledge Distillation (FTKD), which effectively transfers future frame knowledge from an offline teacher model to an online student model. Specifically, we present a future-aware feature reconstruction strategy to encourage the student model to capture future features without strict frame alignment. In addition, we further introduce future-guided logit distillation to leverage the teacher's stable foreground and background context. FTKD is applied to two high-performing 3D object detection baselines, achieving up to 1.3 mAP and 1.3 NDS gains on the nuScenes dataset, as well as the most accurate velocity estimation, without increasing inference cost.
\end{abstract}


\section{Introduction}
\label{sec_1_intro}
Camera-based multi-view 3D object detection has attracted much attention in autonomous driving due to its low deployment cost and rich visual information. Recently, bird's-eye-view (BEV) detection achieves promising performance by incorporating temporal information \cite{li2022bevformer, yang2023bevformer, huang2022bevdet4d, li2023bevdepth, liu2023petrv2, park2022time, han2023exploring, lin2023sparse4d, lin2022sparse4d}. Furthermore, several offline models \cite{liu2023sparsebev, wang2023exploring, lin2023sparse4dv3} introduce future frames through parallel temporal fusion to further boost accuracy, which aids in detecting small or occluded objects. However, online detection lacks access to future frames, making it challenging to effectively utilize this knowledge.

\begin{figure}[t]
\begin{center}
\includegraphics[width=\linewidth]{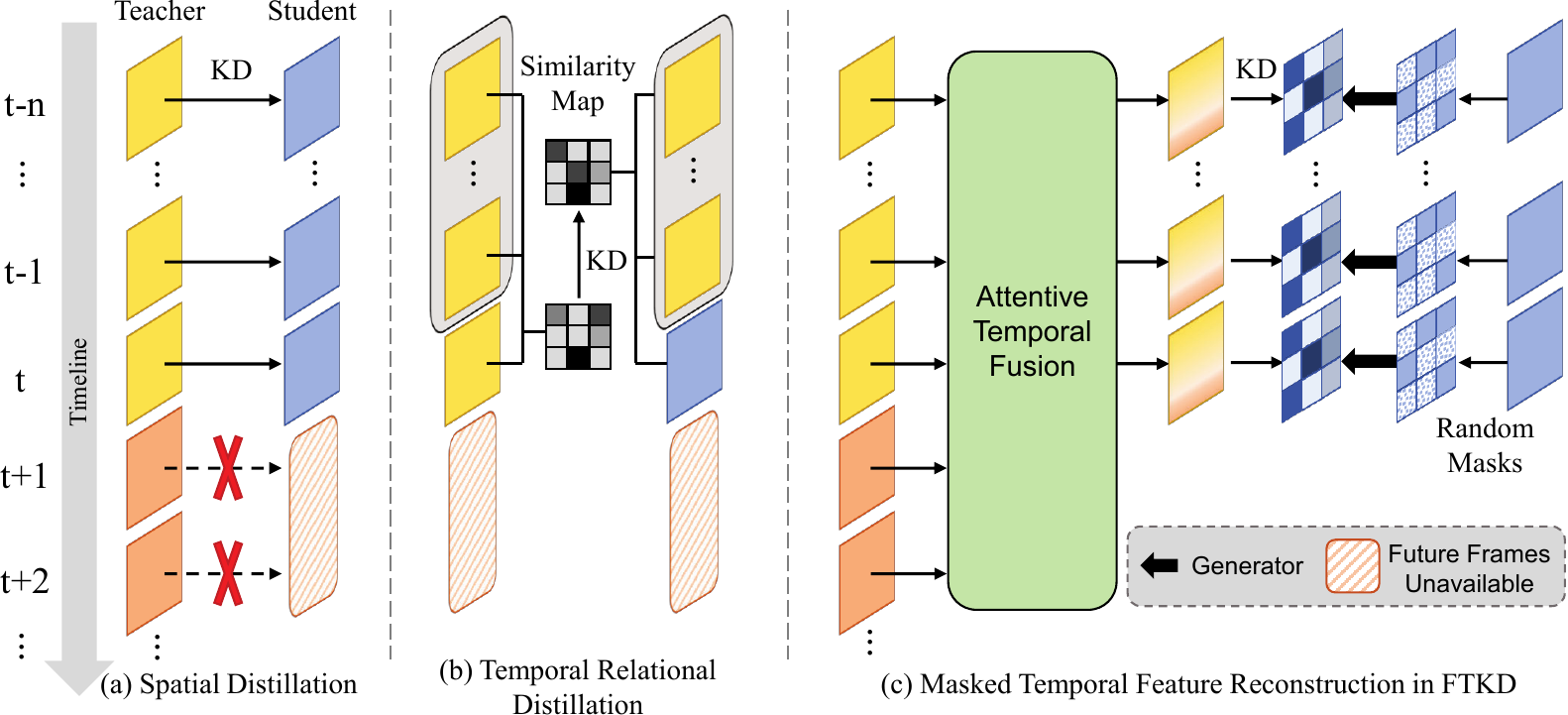}
\end{center}
\caption{Illustration of using future frames in feature distillation. (a) Spatial feature distillation requires strict alignment of input frames between the teacher and student models, preventing the use of future frame information. (b) Temporal relational distillation focuses on inter-frame relational knowledge but overlooks future frames. (c) In FTKD, information from future frames is aggregated temporally and used as the reconstruction objective for student's masked feature, facilitating effective learning of future knowledge.}
\label{fig:comp_kd}
\end{figure}

While knowledge distillation (KD) can be an appealing technique for transferring knowledge from an offline teacher model to an online student model, existing KD methods for 3D object detection still suffer from three key limitations. (i) Most KD methods \cite{yang2022masked, zeng2023distilling, zhou2023unidistill} primarily concentrate on spatial feature distillation, which fail to effectively exploit the teacher's future knowledge due to strict input frame alignment requirements (see Fig. \ref{fig:comp_kd}(a)). (ii) Temporal distillation methods \cite{jang2023stxd, wang2023distillbev} neglect valuable information from future frames, as shown in Fig. \ref{fig:comp_kd}(b). (iii) They apply to weak baselines or are based on dense BEV representation \cite{li2022bevformer, yang2023bevformer, li2023bevdepth, huang2022bevdet4d, park2022time}. Dense BEV methods suffer from increased latency with more input frames, which hinders real-world applicability and deployment.

Moreover, the selection of an appropriate offline teacher model necessitates careful consideration of several factors, including the domain gap across modalities, alignment with future frame information, and the consistency of temporal feature representations. To this end, the teacher model should meet three key criteria: using a camera-based modality to ensure domain alignment, adopting a parallel temporal fusion strategy to effectively integrate temporal cues, and employing a sparse query representation to maintain consistency with the student model's feature representation.

To address the aforementioned problems, we propose a sparse query-based framework, Future Temporal Knowledge Distillation (FTKD). FTKD overcomes the limitations of strict frame alignment in spatial feature distillation. To achieve this, we introduce a future-aware feature reconstruction mechanism that enables the online student model to distill knowledge from future frames, as illustrated in Fig. \ref{fig:comp_kd}(c). The online student model reconstructs complete feature representations from its partial features, guided by a teacher model. This enhances the student's representational capacity by integrating future contextual information. Specifically, we first construct the reconstruction objective by aggregating teacher features that contain future information. Then, we introduce random masks to the student features and generate new features using an adaptive generator. Finally, the student features are reconstructed based on the defined reconstruction objective. This design allows the student detector to incorporate long-term temporal (both historical and future) knowledge without increasing inference overhead.

In addition, since the ground truth contains few foreground bounding boxes, a large portion of the final predictions are assigned to background queries. However, existing KD methods \cite{jang2023stxd, chen2022bevdistill} largely overlook the informative cues embedded in these background queries. Benefiting from stable training with access to future frames, the teacher model can provide more accurate guidance for both foreground and background context. Based on this observation, we propose a future-guided logit distillation (FLD) strategy. Since the order of queries from the teacher and student models is not aligned during distillation, we employ the Hungarian algorithm \cite{kuhn1955hungarian} to establish one-to-one matching between them. This strategy ensures that both foreground and background cues are effectively leveraged, addressing the limitations of prior methods that ignore background supervision. Similarly, the order of student's reconstructed features is also adjusted accordingly before computing the loss. By integrating feature-level and logit-level distillation, the proposed FTKD enables the online student model to effectively learn from future knowledge, while achieving a favorable trade-off between accuracy and efficiency.

In summary, our contributions can be described as
\begin{itemize}
\item  We propose Future Temporal Knowledge Distillation, a camera-only framework that enables online learning from future frames, balancing accuracy and efficiency.
\item We introduce future-aware feature reconstruction and future-guided logit distillation strategies that eliminate the constraint of strict frame alignment and effectively utilize stable background information.
\item The proposed KD method is applied to two high-performing baseline models, and experimental results on the nuScenes dataset validate the effectiveness of FTKD. From the qualitative results, we also improve the detection of occluded and distant objects.
\end{itemize}

\section{Related Work}
\subsection{Camera-based 3D Object Detection}

Recently, camera-based 3D object detection has achieved significant success with bird's eye view (BEV) representation. In terms of representation, 3D object detection is generally categorized into dense BEV-based approaches and sparse query-based approaches. One line of dense BEV-based methods (e.g., LSS \cite{philion2020lift}) attempts to transform multi-view 2D image features into 3D space based on depth estimation. Building upon LSS, the BEVDet series \cite{huang2021bevdet, huang2022bevdet4d} further elevate performance by introducing data augmentation and temporal fusion. Different from the 2D-to-3D back-projection methods, BEVFormer \cite{li2022bevformer} constructs a dense BEV space to sample multi-view 2D features using a deformable attention mechanism. However, dense BEV methods incur substantial computational overhead when modeling temporal information. As a result, inspired by DETR \cite{carion2020end}, sparse query-based approaches are proposed. For instance, DETR3D \cite{wang2022detr3d} initializes a set of 3D queries to explore a sparse BEV representation. However, the performance is compromised if projecting 3D query points to 2D space for sampling camera features with a fixed local receptive field. To address this issue, the PETR series \cite{liu2022petr, liu2023petrv2} leverage global attention to expand the receptive field. Despite the remarkable progress achieved by the aforementioned approaches, they still entail substantial computational burdens. Consequently, SparseBEV presents a fully sparse detector with a scale-adaptive receptive field for a better trade-off between accuracy and speed.

\subsection{Temporal Modeling}
Integrating long-term temporal knowledge is crucial in autonomous driving. Mainstream temporal fusion mechanisms can be divided into parallel temporal fusion \cite{yang2023bevformer, liu2023petrv2, huang2022bevdet4d, li2023bevdepth, park2022time, liu2023sparsebev} and sequential temporal fusion \cite{li2022bevformer, wang2023exploring, han2023exploring, lin2023sparse4d}. Early works \cite{liu2023petrv2, huang2022bevdet4d, li2023bevdepth} fuse short-term memory (2-4 frames), but their performance is not satisfactory. To explore long-term temporal fusion, SOLOFusion \cite{park2022time} utilizes 17 frames and achieves outstanding performance. Nevertheless, parallel temporal fusion methods commonly grapple with the challenge of balancing accuracy and efficiency. To alleviate this problem, \cite{wang2023exploring, han2023exploring, lin2023sparse4d} carry out sequential temporal fusion instead. They propagate historical features into the current timestamp, largely accelerating the inference speed while maintaining excellent performance. However, sequential temporal fusion is limited to past frames, precluding the use of future frames. Furthermore, existing online temporal 3D object detection methods cannot leverage future information. This paper employs a sparse query-based teacher model with parallel temporal fusion to integrate information from future frames. To ensure consistent sparse representations and a balance between accuracy and efficiency, we select two high-performing sparse query-based student models. The two student models \cite{liu2023sparsebev, wang2023exploring} adopt parallel and sequential temporal fusion paradigms, respectively, demonstrating the generalizability of the proposed method across different architectures.

\begin{figure*}[t]
\begin{center}
\includegraphics[width=1.0\linewidth]{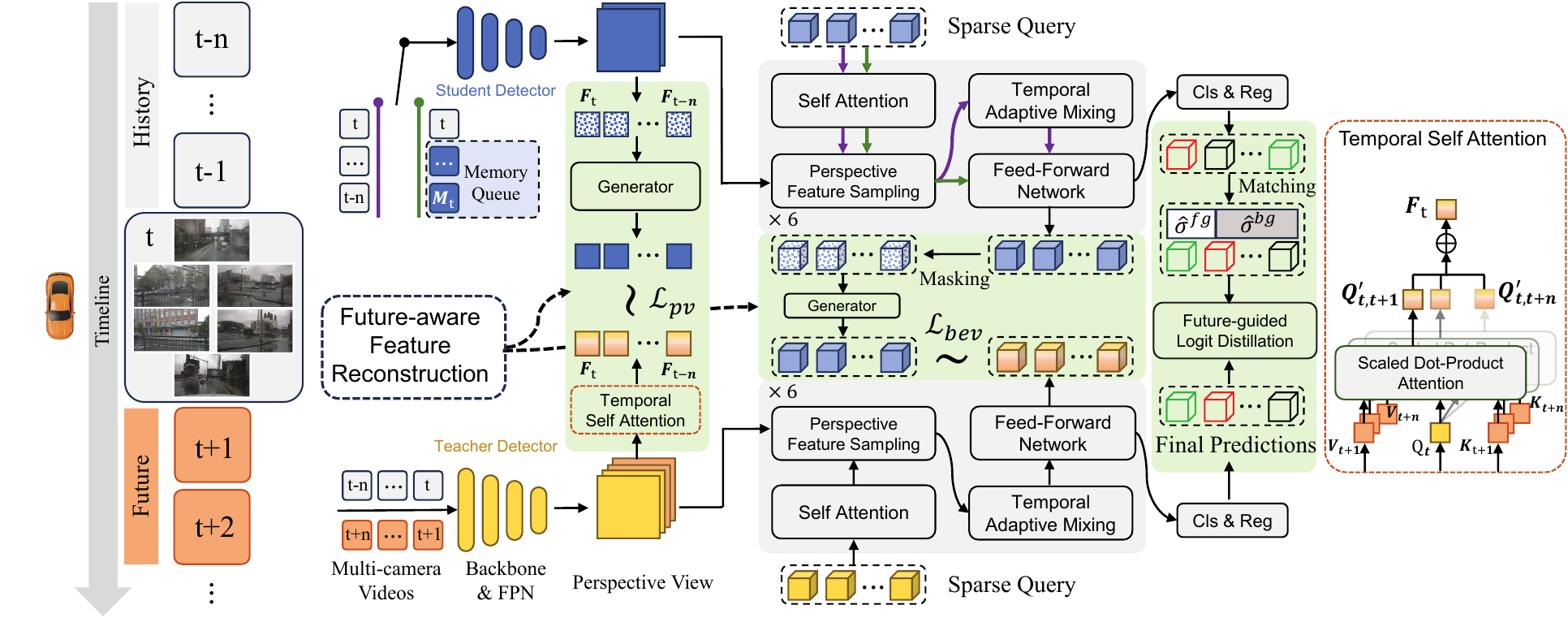}
\end{center}
\caption{Overall framework of Future Temporal Knowledge Distillation (FTKD). FTKD consists of two core distillation components: future-aware feature reconstruction (FFR) and future-guided logit distillation (FLD), which facilitate the transfer of future knowledge from the offline teacher to the online student model. Specifically, FFR conducts masked feature reconstruction on perspective features and sparse BEV query features, while FLD guides the student in capturing both foreground and background cues embedded in the sparse queries.}
\label{fig:framework}
\end{figure*}

\subsection{Knowledge Distillation for Object Detection}
Applying knowledge distillation \cite{hinton2015distilling} on 2D object detection is a popular topic. Different from distilling global features \cite{chen2017learning}, several works emphasize the importance of region selection based on bounding boxes \cite{wang2019distilling, dai2021general, guo2021distilling} and the application of attentive masks on features \cite{zhang2020improve, huang2022masked} to mitigate noise interference. FGD \cite{yang2022focal} effectively combines both strategies, resulting in further performance enhancements. From another perspective, masked feature reconstruction has proved its effectiveness. Inspired by masked image modeling, MGD \cite{yang2022masked} generates random masks on student features, and then reconstructs them under the guidance of teacher features. \cite{huang2022masked} generates attentive masks instead of random masks for feature reconstruction. DETRDistill \cite{chang2023detrdistill} is specifically designed for DETR-families, incorporating both feature-level and logit-level distillation.

For 3D object detection, most methods \cite{zhou2023unidistill, klingner2023x3kd, liu2023geomim, li2022unifying, kim2024labeldistill, huang2023leveraging} focus on cross-modality knowledge distillation, aiming to transfer LiDAR-based features to camera-based features. These methods improve performance by using LiDAR's real-world modeling, but aligning modalities remains challenging, hindering heterogeneous problem processing. To this end, FD3D first proposes camera-only distillation to reconstruct focal knowledge from imperfect teachers. Existing methods largely focus on spatial distillation, leaving temporal distillation underexplored. Although STXD investigates inter-frame relations and DistillBEV distills fused spatiotemporal features, both approaches largely overlook future frame information. In this paper, we propose Future Temporal Knowledge Distillation, effectively transferring future frame information to the online student model, thereby improving performance without introducing any additional inference overhead.


\section{Method}

\subsection{Preliminary: Sparse BEV Models}
Recently, two camera-based 3D detectors, SparseBEV \cite{liu2023sparsebev} and StreamPETR \cite{wang2023exploring}, have shown remarkable performance and high efficiency, both employing sparse BEV query representation. A sparse query is defined as a nine-tuple $Q=(x, y, z, w, l, h, \theta, v_x, v_y)$, where $(x, y, z)$ denotes the query's coordinate in the BEV space, while $w, l, h$ represent its width, length, and height, respectively. $\theta$ and $(v_x, v_y)$ indicate query's rotation and velocity. The query set consists of $N_q$ queries, each associated with $C$-dim features. The two models share a similar detection pipeline but differ in temporal fusion strategy. The query features are first passed through self-attention, followed by sampling from the feature pyramid network (FPN) feature maps to extract semantic information, and finally decoded into the final predictions $\hat{y}=\{\hat{c}, \hat{b}\}$ (categories and bounding boxes) through a decoder layer. Following DETR, the predictions are matched to ground truth (GT) using the Hungarian algorithm for optimal bipartite assignment $\hat{\sigma}$. Finally, the classification and regression losses are computed between the optimally matched predictions $\hat{y}_{\hat{\sigma}}$ and the GT.

SparseBEV employs parallel temporal fusion, aligning with the criteria for an effective teacher model as discussed in Sec. 1. Therefore, we adopt SparseBEV as both the teacher and one of the student models in this paper. In contrast, StreamPETR adopts sequential temporal fusion to accelerate inference, making it a suitable student model to further validate the generalizability of our proposed method.

\subsection{Overall Framework}
Incorporating future frames into temporal modeling can provide richer motion information, thereby improving the detection and velocity estimation performance for dynamic objects. However, online models do not have access to future frames. To address this limitation, we propose Future Temporal Knowledge Distillation (FTKD), as illustrated in Fig. \ref{fig:framework}, which transfers future knowledge from an offline teacher model to an online student model. FTKD consists of two key components: future-aware feature reconstruction and future-guided logit distillation. During the distillation stage, we freeze the teacher detector and retain the original student architecture by removing any extra auxiliary layers post-training. Based on this, there is no additional computational overhead during the inference.

\subsection{Future-aware Feature Reconstruction}
To ensure perceptual consistency, both perspective view (PV) and sparse query features require future-aware feature reconstruction. A key challenge is bridging the gap between the teacher and student models' input frame requirements. This necessitates effectively fusing the teacher's extensive temporal knowledge, encompassing both past and future information. We address this using distinct strategies for PV and sparse query features. Formally, let $\textbf{F}^{T}$ and $\textbf{F}^{S}$ denote the PV or sparse query features from the teacher and student model, conforming to $\textbf{F}^{T}=\{ \{\textbf{F}_{i}^{T\_his}\}^{M^{his}}_{i=1}, \{\textbf{F}^{T\_cur}\}, \{\textbf{F}_{i}^{T\_fut}\}^{M^{fut}}_{i=1} \}$ and $\textbf{F}^{S}=\{ \{\textbf{F}_{i}^{S\_his}\}^{N^{his}}_{i=1}, \{\textbf{F}^{S\_cur}\}\}$, where ${M^{his}}$ and ${M^{fut}}$ indicate the number of historical and future frames of teacher model. $M={M^{his}}+1+{M^{fut}}$ and $N={N^{his}+1}$ are the total numbers of input frames for the teacher and student, respectively. In general, $M^{his}$ is equal to $N^{his}$. 

\noindent\textbf{Temporal self-attention on PV features.} To capture high-level semantic information from future frames while reducing computational overhead, we apply temporal self-attention (TSA) to the final-layer features of the FPN. $\textbf{F}_{pv}^{T_0}=\{ \{\textbf{F}_{pv,i}^{T\_his}\}^{M^{his}}_{i=1}, \{\textbf{F}_{pv}^{T\_cur}\} \}$ is used as the query, while $\{\textbf{F}_{pv,i}^{T\_fut}\}^{M^{fut}}_{i=1}$ serves as the key and value to extract informative cues from future frames. Therefore, the aggregation of teacher temporal knowledge can be formulated as

\begin{equation}
    \textbf{F}^{T\_agg}_{pv,i} = \sum_{j=1}^{M^{fut}} \text{TSA}(\textbf{F}^{T_0}_{pv,i}, \textbf{F}^{T\_fut}_{pv,j}, \textbf{F}^{T\_fut}_{pv,j}),
    \label{eq1_TSA}
\end{equation}
where $i=1,2,...,M^{his}+1$ and TSA is based on the scaled dot-product attention operation \cite{vaswani2017attention}.

\noindent\textbf{Temporal adaptive mixing on sparse query features.} Since AdaMixer \cite{gao2022adamixer} and SparseBEV propose an efficient and adaptive sparse query decoding mechanism, we reuse it to fuse temporal (both historical and future) query features and finally obtain $\textbf{F}^{T\_agg}_{bev}$.

Once the reconstruction objective is defined, we perform masked reconstruction on the student features. Random mask is generated on $\textbf{F}^{S}$ with mask ratio $\lambda$, which can be formulated as

\begin{equation}
    M_{k,i} = \begin{cases}
    0, \text{if} \ R_{k,i} < \lambda \\
    1, \text{otherwise}
    \end{cases},
\end{equation}
where $R_{k,i}$ is a random number in $(0,1)$ and $k$ indicates the index of query or pixel. $i$ denotes the $i$-th frame. Subsequently, we recover masked student features using a generation layer $\mathcal{G}$:

\begin{equation}
    \hat{\textbf{F}}^{S} = \mathcal{G}(\textbf{F}^{S} \cdot M).
    \label{eq3_gen}
\end{equation}
Since the PV features and sparse query features of the student model have different dimensions, we design separate generation layers for them accordingly. For PV features, $\mathcal{G}$ is composed of two 2D $3 \times 3$ convolutional layers and one $ReLU$ layer, whereas for sparse query features, $\mathcal{G}$ consists of a feed-forward network (FFN) followed by a layer normalization. Then the generated student features $\hat{\textbf{F}}^{S}$ are reconstructed under the supervision of the teacher's temporally aggregated features $\textbf{F}^{T\_agg}$ using Mean Squared Error (MSE). Thus, the temporal reconstruction loss for PV features can be represented as

\begin{equation}
    \mathcal{L}_{pv} = \frac{1}{n} \sum_{i=1}^{N} \sum_{l=1}^{L} \sum_{c=1}^{C} \Vert \hat{\textbf{F}}^{S}_{pv,i,l,c} - \textbf{F}^{T\_agg}_{pv,i,l,c}\Vert^2_2,
    \label{eq4_pvloss}
\end{equation}
where L denotes the product of the height and width of the feature map and $n=N \times L \times C$.

Similarly, the reconstruction loss for sparse query features can be formulated as

\begin{equation}
    \mathcal{L}_{bev} = \frac{1}{n} \sum_{i=1}^{N} \sum_{q=1}^{N_q} \sum_{c=1}^{C} \Vert \hat{\textbf{F}}^{S}_{bev,i,\hat{\sigma}_q,c} - \textbf{F}^{T\_agg}_{bev,i,q,c}\Vert^2_2,
    \label{eq5_bevloss}
\end{equation}
where $n=N \times N_q \times C$. $\hat{\sigma}_q$ represents the optimal permutation of $N_q$ elements obtained by applying the Hungarian algorithm to match the teacher's final predictions with those of the student.

\begin{table*}[t]
\begin{center}
\resizebox{\linewidth}{!}{
\begin{tabular}{l|c|cc|ccccc|c}
\toprule[1pt]
Method          & Frames & NDS $\uparrow$ & mAP $\uparrow$  & mATE $\downarrow$ & mASE $\downarrow$ & mAOE $\downarrow$ & mAVE $\downarrow$ & mAAE $\downarrow$ & FPS $\uparrow$ \\  \midrule
\multicolumn{10}{c}{Results without distillation schemes} \\ \midrule
BEVDet4D $\dag$ \cite{huang2022bevdet4d}  & 2 & 45.7 & 32.2 & 0.703 & 0.278 & 0.495 & 0.354 & 0.206 & 30.7 \\
PETRv2 \cite{liu2023petrv2}  & 2 & 45.6 & 34.9 & 0.700 & 0.275 & 0.580 & 0.437 & 0.187 & - \\
SOLOFusion $\dag$ \cite{park2022time} & 16+1 & 53.4 & 42.7 & 0.567 & 0.274 & 0.511 & 0.252 & 0.181 & 15.7* \\ \midrule
VideoBEV $\dag$ \cite{han2023exploring} & 8 & 53.5 & 42.2 & 0.564 & 0.276 & 0.440 & 0.286 & 0.198 & - \\
Sparse4Dv2 \cite{lin2023sparse4d} & - & 53.9 & 43.9 & 0.598 & 0.270 & 0.475 & 0.282 & 0.179 & 17.3 \\ \midrule
\multicolumn{10}{c}{Results with distillation schemes} \\ \midrule

T:SparseBEV-R101 \cite{liu2023sparsebev} & 15      & 63.8 & 55.1 & 0.493 & 0.265 & 0.275 & 0.163 & 0.181 & 3.1 \\
S:SparseBEV-R50 & 8 & 55.5 & 44.7 & 0.585 & 0.271 & 0.391 & 0.251 & 0.188 & 20.2 \\ 
\hspace{1em} +MGD \cite{yang2022masked} & 8      & 55.1 {\footnotesize($\downarrow$0.4)} & 44.8 {\footnotesize($\uparrow$0.1)}& 0.591 & 0.270 & 0.425 & 0.248 & 0.192 & 20.2 \\
\hspace{1em} +CWD \cite{shu2021channel} & 8      & 55.1 {\footnotesize($\downarrow$0.4)} & 44.6 {\footnotesize($\downarrow$0.1)} & 0.591 & 0.270 & 0.408 & 0.251 & 0.190 & 20.2 \\
\hspace{1em} +FD3D \cite{zeng2023distilling} & 8   & 55.0 {\footnotesize($\downarrow$0.5)} & 44.6 {\footnotesize($\downarrow$0.1)} & 0.598 & 0.270 & 0.423 & 0.251 & 0.187 & 20.2 \\
\hspace{1em} +STXD \cite{jang2023stxd} & 8 & 55.6 {\footnotesize($\uparrow$0.1)} & 45.0 {\footnotesize($\uparrow$0.3)} & 0.588 & 0.271 & 0.398 & 0.247 & 0.187 & 20.2 \\
\rowcolor{gray!40}
\hspace{1em} +FTKD (ours) & 8     & \textbf{56.5} {\footnotesize($\uparrow$1.0)} & \textbf{46.0} {\footnotesize($\uparrow$1.3)} & \textbf{0.579} & \textbf{0.270} & \textbf{0.372} & \textbf{0.234} & \textbf{0.179} & 20.2 \\

S:StreamPETR-R50 \cite{wang2023exploring} & 8      & 55.0 & 45.0 & 0.613 & 0.267 & 0.413 & 0.265 & 0.196 & \textbf{33.9} \\
\hspace{1em} +MGD \cite{yang2022masked} & 8      & 55.1 {\footnotesize($\uparrow$0.1)} & 45.0 \footnotesize($\uparrow$0.0) & 0.618 & 0.269 & 0.402 & 0.260 & 0.194 & \textbf{33.9} \\
\hspace{1em} +CWD \cite{shu2021channel} & 8      & 54.8 {\footnotesize($\downarrow$0.2)} & 44.8 {\footnotesize($\downarrow$0.2)} & 0.610 & 0.270 & 0.428 & 0.263 & 0.189 & \textbf{33.9} \\
\hspace{1em} +FD3D \cite{zeng2023distilling} & 8  & 55.4 {\footnotesize($\uparrow$0.4)} & 45.3 {\footnotesize($\uparrow$0.3)} & 0.600 & 0.268 & 0.405 & 0.259 & 0.190 & \textbf{33.9} \\
\hspace{1em} +STXD \cite{jang2023stxd} & 8 & 55.6 {\footnotesize($\uparrow$0.6)} & 45.5 {\footnotesize($\uparrow$0.5)} & 0.597 & 0.269 & 0.411 & 0.254 & 0.184 & \textbf{33.9} \\
\rowcolor{gray!40}
\hspace{1em} +FTKD (ours) & 8  & \textbf{56.3} {\footnotesize($\uparrow$1.3)} & \textbf{46.3} {\footnotesize($\uparrow$1.3)} & \textbf{0.589} & \textbf{0.268} & \textbf{0.398} & \textbf{0.252} & \textbf{0.182} & \textbf{33.9} \\ 

\bottomrule[1pt]
\end{tabular}
}
\caption{Comparison on the nuScenes validation set. $\dag$ denotes methods with CBGS \cite{zhu2019class}. Only the teacher model has access to future frames. The student baselines benefit from perspective-view pretraining. FPS is measured on RTX4090 with fp32 without cuda acceleration. * represents inference with fp16. The input size for ResNet101 (R101) and ResNet50 (R50) are 512 $\times$ 1408 and 256 $\times$ 704, respectively.}
\end{center}

\label{tab:main_exp}
\end{table*}

\subsection{Future-guided Logit Distillation}

Logit distillation has been widely adopted in KD. 
In cross-modal distillation \cite{chen2022bevdistill, jang2023stxd}, these methods typically assume that the teacher's final predictions vary in importance. Therefore, they assign quality scores as weights to the teacher predictions. However, such approaches often neglect background information. To address this limitation, inspired by DETRDistill \cite{chang2023detrdistill}, we introduce FLD.

With the guidance of future knowledge, the teacher model is well optimized and can consistently provide a large number of true negatives. Therefore, we apply the Hungarian algorithm to perform bipartite matching between the teacher's predictions $\hat{y}^T$ and the student's predictions $\hat{y}^S$. This yields the optimal permutation of both foreground and background samples, denoted as $\hat{\sigma}^{fg}$ and $\hat{\sigma}^{bg}$, respectively. Then the logit distillation can be formulated as 

\begin{equation}
    \mathcal{L}_{logits} = \sum_{q=1}^{N_q} \alpha \mathcal{L}_{cls}(\hat{c}^{S}_{\hat{\sigma}_q}, \hat{c}^{T}_q) + \beta \mathcal{L}_{bbx}(\hat{b}^{S}_{\hat{\sigma}_q}, \hat{b}^{T}_q),
\label{eq6_logitloss}
\end{equation}
where $\hat{\sigma}=\{\hat{\sigma}^{fg},\hat{\sigma}^{bg}\}$. $\mathcal{L}_{cls}$ and $\mathcal{L}_{bbx}$ is FocalLoss \cite{lin2017focal} and L1 loss, respectively. $\alpha$ and $\beta$ are weights to balance the logit KD loss (set to 2.0 and 0.25 by default).

\subsection{Overall Distillation Loss}
Finally, the overall KD loss for the online student model is formulated by integrating Eq. \ref{eq4_pvloss}, Eq. \ref{eq5_bevloss}, and Eq. \ref{eq6_logitloss}:

\begin{equation}
    \mathcal{L}_{KD} = \lambda_1 \mathcal{L}_{pv} + \lambda_2 \mathcal{L}_{bev} + \lambda_3 \mathcal{L}_{logits},
\end{equation}
where $\lambda_1$, $\lambda_2$ and $\lambda_3$ are loss weights to balance the KD losses. In summary, we train the student model with original classification and regression loss as well as KD loss $\mathcal{L}_{KD}$.

\section{Experiments}

\begin{table}[t]
\centering
\begin{tabular}{c|c|cc}
\toprule
Model & Future Frame     & NDS $\uparrow$ & mAP $\uparrow$  \\ \midrule
\multirow{2}{*}{SparseBEV-R50} & \usym{2613}            & 55.5 & 45.0  \\

& {\checkmark} & \textbf{56.5} & \textbf{46.0}  \\ \midrule 
\multirow{2}{*}{StreamPETR-R50} & \usym{2613}            & 55.2 & 45.1  \\

& {\checkmark} & \textbf{56.3} & \textbf{46.3}  \\
\bottomrule[1pt]
\end{tabular}
\caption{Ablation on the utilization of future frames in KD. We use SparseBEV-R101 as the teacher model, with 8 input frames (7 historical frames and 1 current frame).}
\label{tab5:ab_future}
\end{table}

\subsection{Datasets and Metrics}
A large-scale surround-view autonomous driving benchmark, nuScenes \cite{caesar2020nuscenes}, is utilized to evaluate our approach. It comprises 700/150/150 scenes for training/validation/testing. Each scene spans approximately 20 seconds, with annotations available for key frames at 0.5s intervals. For 3D object detection, it includes 1.4M 3D bounding boxes across 10 categories. Following the official evaluation metrics, we report nuScenes detection score (NDS), mean Average Precision (mAP), and five true positive (TP) metrics, including ATE, ASE, AOE, AVE, and AAE for measuring translation, scale, orientation, velocity, and attributes, respectively. The NDS combines mAP and five TP metrics to provide a comprehensive evaluation score.

\subsection{Implementation Details} 
The experimental results are reported based on 8 A100 GPUs, and FPS measurements are conducted on RTX4090 with fp32. We train all the models with AdamW \cite{loshchilov2017decoupled} optimizer for 24 epochs on SparseBEV and 60 epochs on StreamPETR, using perspective pretraining on nuImage \cite{caesar2020nuscenes}. The initial learning rate is set to $2 \times 10^{-4}$ and is decayed with a cosine annealing policy. The global batch size is fixed to 8. For supervised training, the Hungarian algorithm is used for label assignment. FocalLoss and L1 loss are employed for classification and 3D bounding boxes regression, respectively. We initialize $N_q=900$ queries and set the channel of query features $C=256$. The mask ratio $\lambda$ is fixed to 0.5. Loss weights $\lambda_1$, $\lambda_2$, and $\lambda_3$  are set to $1e^{-3}$, $16$ and 1, respectively. We set the number of frames $M^{his}=M^{fut}=N^{his}=7$ by default. All data preprocessing follows the corresponding baselines.

\subsection{Main Results}
\textbf{Comparison with spatial distillation methods.} Conventional 2D spatial distillation methods \cite{yang2022masked, shu2021channel} struggle to improve temporal 3D object detection, possibly due to their inability to capture temporal dynamics and 3D spatial context. FD3D is the first camera-only KD method that is designed for the spatial domain. We reimplement it for a fairer comparison. Our method outperforms FD3D with improvements of 1.4 mAP and 1.5 NDS on SparseBEV. Note that FD3D requires strictly aligned input frames between the student and teacher models, which prevents the transfer of information from future frames. Hence, only spatial knowledge from past and current frames can be distilled. This limitation, combined with its neglect of background information, likely contributes to its suboptimal performance in temporal 3D object detection.

\begin{table}[t]

\centering
\begin{tabular}{ccc|cc|c}
\toprule[1pt]
\multicolumn{2}{c}{FFR} & \multirow{2}{*}{FLD} & \multirow{2}{*}{NDS $\uparrow$} & \multirow{2}{*}{mAP $\uparrow$} & \multirow{2}{*}{mAVE $\downarrow$} \\ \cline{1-2}
PV   & BEV   &  &  &  &  \\ \midrule
     &       &  & 55.5 & 44.7 & 0.251 \\
\multicolumn{1}{c} {\checkmark} & & & 55.7 & 44.9 & 0.250 \\
 & {\checkmark} &                & 55.8 & 45.2 & 0.243 \\ 
 &       & {\checkmark}          & 55.9 & 45.3 & 0.247 \\ 
 & {\checkmark} & {\checkmark}   & 56.3 & 45.6 & 0.235 \\ 
\multicolumn{1}{c} {\checkmark} & {\checkmark} &  & 55.9 & 45.4 &  0.243 \\
\rowcolor{gray!40}
\multicolumn{1}{c} {\checkmark} & {\checkmark} & {\checkmark} & \textbf{56.5} & \textbf{46.0} & \textbf{0.234} \\
\bottomrule[1pt]
\end{tabular}
\caption{Effectiveness of loss components. FFR indicates future-aware feature reconstruction, which is applied on perspective view (PV) and sparse bird's-eye-view (BEV) query, respectively. FLD denotes future-guided logit distillation.}
\label{tab2:loss_comp}
\end{table}

\noindent\textbf{Comparison with temporal distillation method.} Although STXD is a cross-modal KD method, its temporal distillation component is highly relevant to our temporal student detectors. Therefore, we reproduce STXD on both SparseBEV and StreamPETR. For a fair comparison, the reproduced STXD is also adapted to explore relational knowledge from future frames. Our FTKD outperforms STXD by 0.9 NDS and 1.0 mAP on SparseBEV, and by 0.7 NDS and 0.8 mAP on StreamPETR. The insufficient utilization of future knowledge and background information may be the main factors limiting the performance of STXD.

\subsection{Ablation Study}

\noindent\textbf{Ablation on future frames.} We first validate the effectiveness of incorporating future frames in knowledge distillation. While high-performance 3D object detectors have demonstrated strong capabilities in capturing information from historical frames, our results show that future cues can further benefit online student models, demonstrating the strength of our approach.

\begin{figure}[t]
\begin{center}
\includegraphics[width=\linewidth]{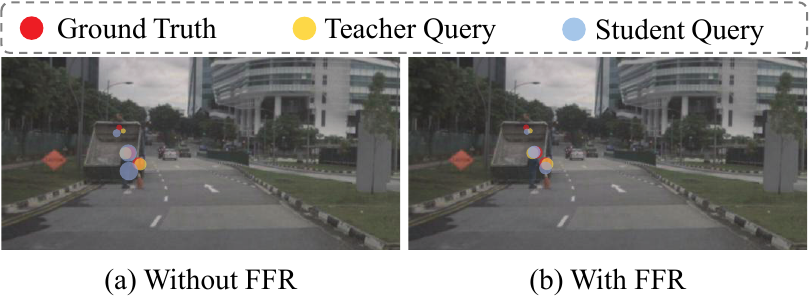}
\end{center}
\caption{Visualization of sparse queries (a) with and (b) without future-aware feature reconstruction (FFR). Larger points denote shallower depth. It is evident that, with FFR, the sparse queries are more aligned with the ground truth.}
\label{fig:ffr}
\end{figure}

\noindent\textbf{Effects of loss components.} In Table \ref{tab2:loss_comp}, we evaluate the effects of each loss component on NDS, mAP, and mAVE. Notably, employing FFR solely on BEV yields higher accuracy (NDS: $\uparrow 0.3$, mAP: $\uparrow 0.5$) than the other two loss terms. Moreover, the other two loss terms can also improve 0.2-0.4 NDS. When applying FLD and FFR (on sparse BEV query), we observe significant enhancements in NDS and mAP, reaching 56.3 and 45.6, respectively. This emphasizes the efficacy of FFR and FLD, providing the model with future knowledge and background information. As a result, the combination of these two types of distillation can lead to certain performance gains. We also provide a visualization of the sparse queries in Fig. \ref{fig:ffr}, revealing that the teacher model effectively guides their feature reconstruction.

\begin{table}[t]

\centering
\begin{tabular}{c|c|cc|c}
\toprule[1pt]
Location &Mask ratio & NDS $\uparrow$ & mAP $\uparrow$ & mAVE $\downarrow$ \\ \midrule
\multirow{5}{*}{BEV} & 0.4 & 55.4 & 44.8 & 0.251 \\
    & \cellcolor{gray!40}0.5 & \cellcolor{gray!40}\textbf{55.8} & \cellcolor{gray!40}\textbf{45.2} & \cellcolor{gray!40}\textbf{0.243} \\
                    & 0.6 & 55.5 & 45.1 & 0.247 \\
                    & 0.75 & 54.9 & 45.1 & 0.253 \\
                    & 0.9 & 54.8 & 44.3 & 0.268 \\ \midrule
\multirow{3}{*}{BEV \& PV} & \cellcolor{gray!40}0.5 \& 0.5 & \cellcolor{gray!40}\textbf{55.9} & \cellcolor{gray!40}\textbf{45.4} & \cellcolor{gray!40}\textbf{0.243} \\
                        & 0.5 \& 0.65 & 55.0 & 45.2 & 0.261 \\
                        & 0.5 \& 0.75 & 55.2 & 45.1 & 0.255 \\
                    
\bottomrule[1pt]
\end{tabular}
\caption{Ablation of the mask ratio.}
\label{tab3:mask_ratio}
\end{table}

\begin{figure*}[ht]
\begin{center}
\includegraphics[width=0.92\linewidth]{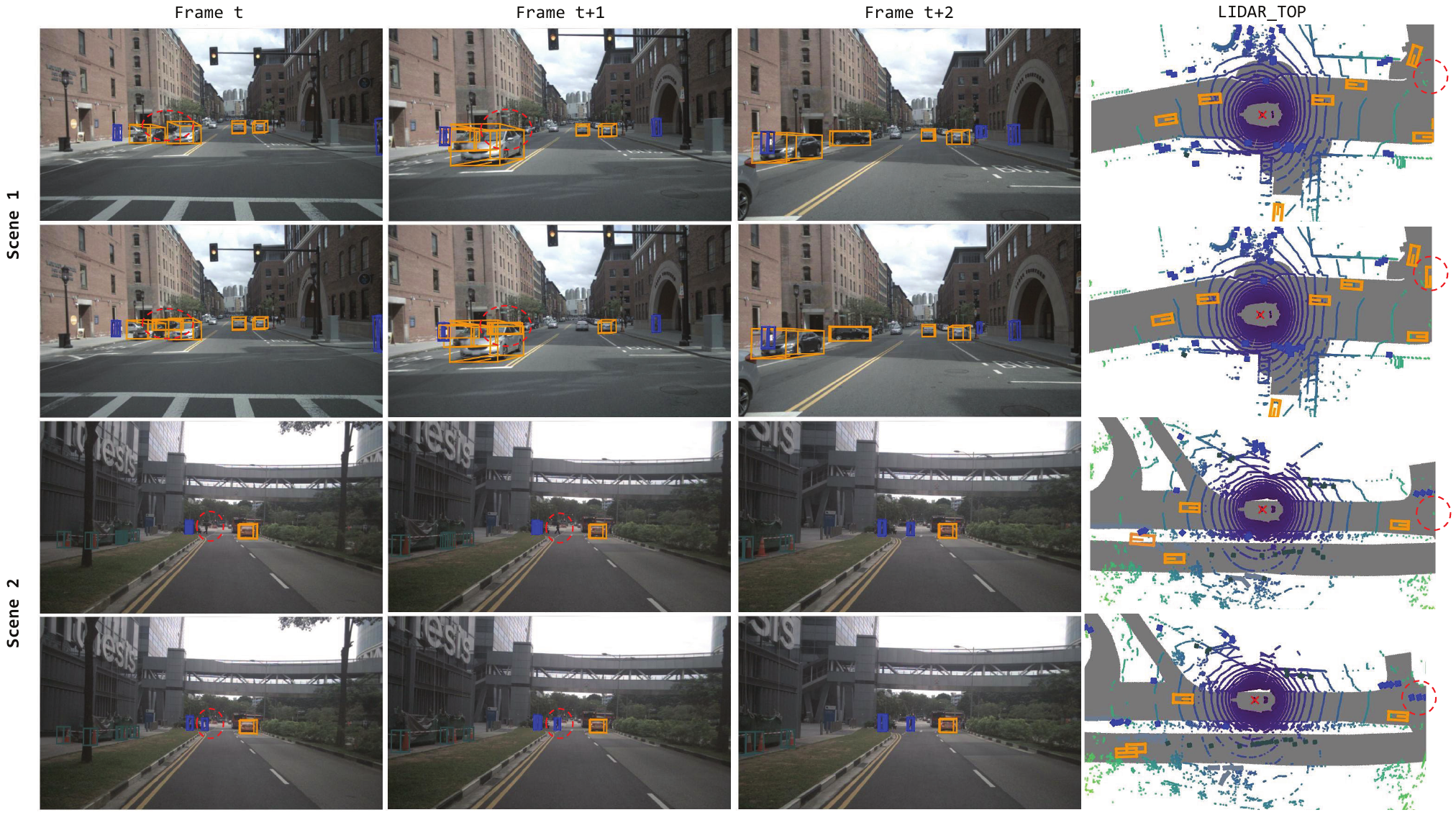}
\end{center}
\caption{Qualitative results over three consecutive frames (front camera) in two scenes. The first and third row show the prediction made by the baseline model, while the second and fourth row demonstrate the predictive results of FTKD. In the last column, the LiDAR point cloud in BEV is display for frame $t+1$, except the last row (for $t+2$) due to the limited BEV distance. FTKD successfully predicts an occluded car merging into the main road and a pedestrian crossing the street in the distance, highlighted by red dotted circles.}
\label{fig:visualization}
\end{figure*}

\noindent\textbf{Impacts of mask ratio.} We investigate various mask ratios for FFR in both sparse BEV query and PV. As shown in Table \ref{tab3:mask_ratio}, we can identify the optimal mask ratio to be 0.5 for both BEV and PV. The core idea of masked feature reconstruction is to use the residual features to reconstruct complete feature maps. A high mask ratio results in a poor representation of residual features, while a low mask ratio simplifies the generator's learning process, enabling shortcuts that lead to local optima. Moreover, a low mAVE demonstrates that temporal feature reconstruction is conducive to focusing on dynamic objects and estimating their velocity.

\begin{table}[t]
\centering
\begin{tabular}{c|cc|c}
\toprule
selections of FLD     & NDS $\uparrow$ & mAP $\uparrow$ & mAVE $\downarrow$ \\ \midrule
foreground            & 55.4 & 44.9 & 0.254 \\
background            & 55.5 & 45.1    & 0.251 \\
\rowcolor{gray!40}
foreground+background & \textbf{55.9} & \textbf{45.3} & \textbf{0.247} \\ 
\bottomrule[1pt]
\end{tabular}
\caption{Ablation on the selections of FLD.}
\label{tab4:ab_fld}
\end{table}

\noindent\textbf{Selections of future-guided logit distillation.} Our ablation study on logit distillation reveals that using only the positive foreground queries yields limited performance improvement, potentially because GT already provides foreground information. However, performance improvements from using only background queries suggest that exploiting these queries is beneficial. The results suggest that future-guided teacher models can enhance student performance by offering both stable foreground and background cues.

\subsection{Qualitative results}
We provide qualitative results to visualize the prediction of the model with and without FTKD, which highlights the importance of future knowledge and the superior performance of the proposed FTKD. Specifically, FTKD exhibits enhanced capabilities in detecting occluded objects and distant targets. As shown in the second row in Fig. \ref{fig:visualization}, an occluded vehicle about to merge onto the main road can be detected earlier. Furthermore, FTKD successfully identifies a distant pedestrian in advance (see the last row in Fig. \ref{fig:visualization}). These examples demonstrate that FTKD can effectively capture knowledge from future frames, contributing to improved detection of occluded and distant objects.


\section{Conclusion}
In this work, we introduce Future Temporal Knowledge Distillation (FTKD), a sparse query-based framework that effectively transfers knowledge from future frames encoded by an offline teacher model to an online student model. FTKD leverages future-aware feature reconstruction to overcome frame alignment constraints in spatial distillation, facilitating effective future temporal knowledge transfer. Additionally, future-guided logit distillation enriches the student model's understanding of background information. Experiments conducted on two strong detectors validate the effectiveness of FTKD and show that it maintains a desirable trade-off between accuracy and efficiency.

\noindent\textbf{Limitations.} FTKD is only validated on the 3D object detection task under the camera modality. For autonomous driving, exploring how to effectively learn future knowledge in multi-modal settings and other 3D perception tasks (e.g., 3D occupancy prediction) remains an important research topic.

\section{Acknowledgments}
This work was supported by the Science and Technology Development Fund of Macau Project 0096/2023/RIA2, and in part by Chinese National Natural Science Foundation Projects 62206276.

\bibliography{aaai2026}

\end{document}